\documentclass{article}
\usepackage{ijcai25}

\usepackage{times}
\usepackage{soul}
\usepackage{url}
\usepackage[hidelinks]{hyperref}
\usepackage[utf8]{inputenc}
\usepackage[small]{caption}
\usepackage{graphicx}
\usepackage{amsmath}
\usepackage{amsthm}
\usepackage{booktabs}
\usepackage{algorithm}
\usepackage{algorithmic}
\usepackage[switch]{lineno}

\usepackage{pifont}
\newcommand\myfootnotestyle[1]{\ifcase#1 \or \ding{182}\or \ding{183}\or
\ding{184}\or \ding{185}\or \ding{186}\or \ding{187}%
\or \ding{188}\or \ding{189}\or \ding{190}\or \ding{191}\else *\fi\relax}
\usepackage{subfig}
\usepackage{multirow}
\usepackage{tcolorbox}
\usepackage{color}
\usepackage{mathtools}
\usepackage{amsfonts}
\usepackage{bbm}
\usepackage{bm}
\usepackage{makecell}

\title{Connecting Giants: Synergistic Knowledge Transfer of Large Multimodal Models for Few-Shot Learning}

\author{
Hao Tang$^{1}$
\and
Shengfeng He$^{2*}$
\And
Jing Qin$^{1}$\\
\affiliations
$^1$Centre for Smart Health, The Hong Kong Polytechnic University\\
$^2$School of Computing and Information Systems, Singapore Management University\\
\emails
\{howard-hao.tang, harry.qin\}@polyu.edu.hk, shengfenghe@smu.edu.sg
}

\begin{document}

\maketitle

\let\thefootnote\relax
\footnotetext{$^*$ Corresponding author.}

\begin{abstract}
Few-shot learning (FSL) addresses the challenge of classifying novel classes with limited training samples. While some methods leverage semantic knowledge from smaller-scale models to mitigate data scarcity, these approaches often introduce noise and bias due to the data’s inherent simplicity. 
In this paper, we propose a novel framework, Synergistic Knowledge Transfer (\textsc{SynTrans}), which effectively transfers diverse and complementary knowledge from large multimodal models to empower the off-the-shelf few-shot learner. 
Specifically, \textsc{SynTrans} employs CLIP as a robust teacher and uses a few-shot vision encoder as a weak student, distilling semantic-aligned visual knowledge via an unsupervised proxy task.
Subsequently, a training-free synergistic knowledge mining module facilitates collaboration among large multimodal models to extract high-quality semantic knowledge. 
Building upon this, a visual-semantic bridging module enables bi-directional knowledge transfer between visual and semantic spaces, transforming explicit visual and implicit semantic knowledge into category-specific classifier weights.
Finally, \textsc{SynTrans} introduces a visual weight generator and a semantic weight reconstructor to adaptively construct optimal multimodal FSL classifiers. 
Experimental results on four FSL datasets demonstrate that \textsc{SynTrans}, even when paired with a simple few-shot vision encoder, significantly outperforms current state-of-the-art methods. 
\end{abstract} 
\section{Introduction}
\label{intro}
Deep learning models have achieved remarkable success in numerous computer vision tasks~\cite{LiTM19}. However, their effectiveness typically relies on deep neural architectures~\cite{HeZRS16} and large-scale training datasets~\cite{russakovsky2015imagenet}, which hinders their applicability in real-world scenarios where annotated data are scarce. In contrast, humans exhibit an exceptional ability to acquire new concepts and recognize categories from only a handful of samples, aided by extensive prior knowledge and contextual understanding~\cite{ralph2017neural}. This gap has motivated researchers to investigate few-shot learning (FSL)~\cite{tang2020blockmix,wu2022information}, where a model classifies query samples into one of $N$ novel classes, each provided with only $K$ labeled examples.

\begin{figure*}[t!]
\begin{center}
\includegraphics[width=0.85\textwidth]{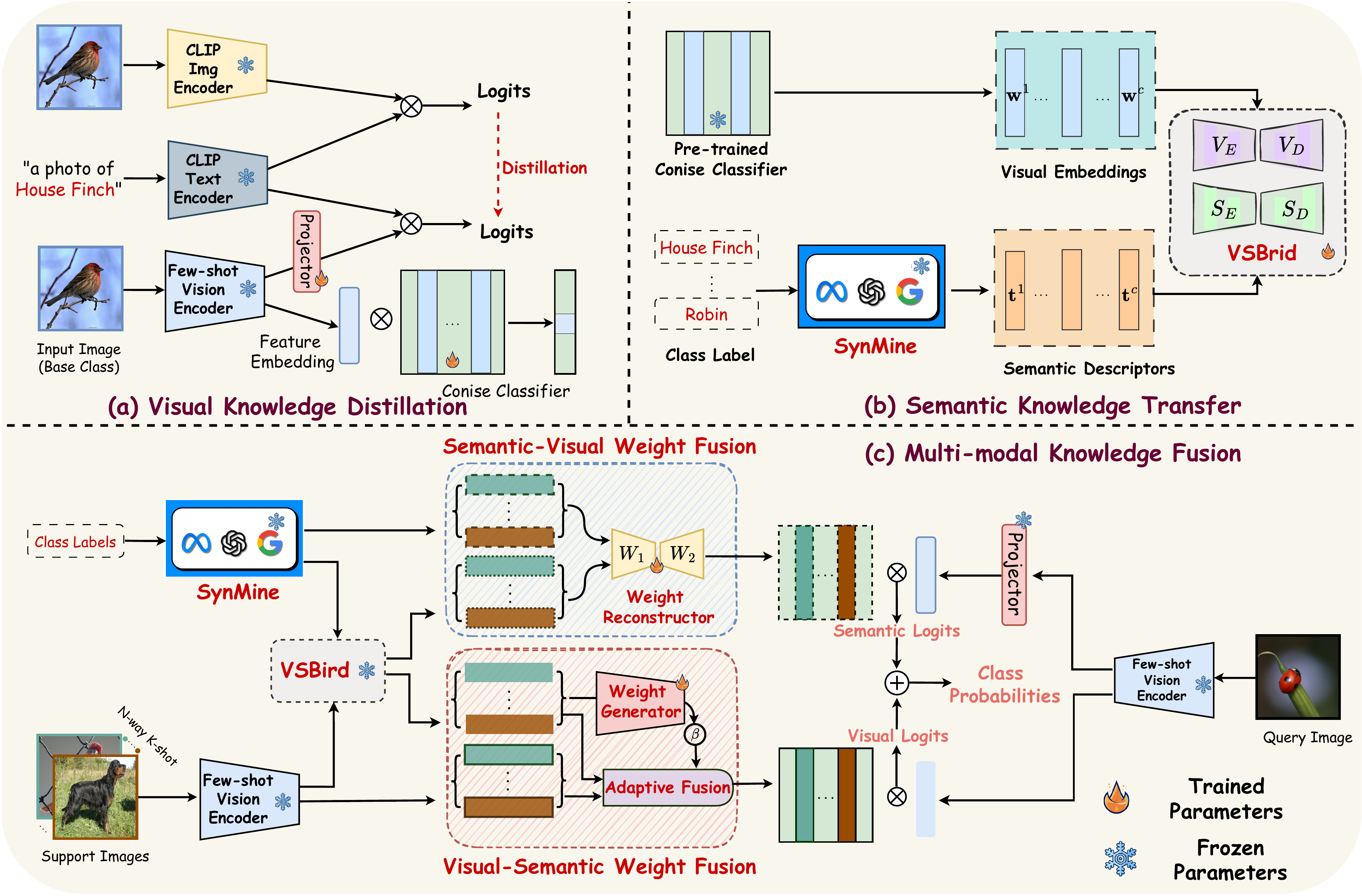}
\end{center}
  \caption{
The pipeline of the proposed \textbf{Syn}ergistic Knowledge \textbf{Trans}fer (\textsc{SynTrans}) framework.
 }
\label{fig:overview}
\end{figure*}

The effectiveness of FSL heavily relies on leveraging prior knowledge to address data scarcity. Existing methods commonly transfer knowledge~\cite{TangYLT22,ZhaTST23} from a disjoint base dataset to novel classes. Early works primarily focused on efficiently exploiting visual prior knowledge, including metric-based~\cite{snell2017prototypical} and optimization-based paradigms~\cite{ravi2016optimization}, both striving to train a base learner capable of rapid adaptation to novel classes with limited training data. While these methods have achieved promising results, there remains a huge gap in comparison to how humans utilize accumulated knowledge and experiences~\cite{YanDZT23,YanTZT24}. As a result, semantic-based methods have emerged to explore various types of semantic knowledge as auxiliary information to improve FSL performance. This semantic knowledge can be obtained either manually (\emph{i.e.,~}attribute annotations) or automatically (\emph{i.e.,~}word vectors). Unfortunately, acquiring attribute annotations requires substantial human effort and may be infeasible for large-scale datasets, while word vectors derived from a single class name tend to be noisy or lack contextual richness. Hence, how to effectively collect and utilize high-quality prior knowledge in FSL is worthy of further investigation.

\textbf{Perceptual filling-in}~\cite{NeumannPH01} is a fundamental characteristic of the human visual system, in which the brain employs prior knowledge and contextual cues to intuitively ``\textit{fill in}'' missing information, resulting in a coherent and comprehensive perception. This phenomenon becomes particularly apparent in scenarios where visual stimuli are limited or partially obscured~\cite{TangLZHT25}, enabling a seamless visual experience despite incomplete data.
Inspired by this phenomenon, we hypothesize that transferring rich external knowledge to the off-the-shelf few-shot learner can further improve performance. Recently, Large Multimodal Models (LMMs)~\cite{Ouyang0JAWMZASR22,RadfordKHRGASAM21} containing abundant implicit knowledge have emerged as powerful repositories of external knowledge in various domains. These models encompass extensive understandings of the visual world, language, and inter-entity relationships, encapsulating diverse knowledge and information. A natural question thus arises: \textbf{can we replicate this human-like cognitive process by connecting the rich, multimodal knowledge of these ``giants'' to compensate for incomplete visual data?}

In this paper, we propose a Synergistic Knowledge Transfer (\textsc{SynTrans}) framework to harness the extensive knowledge embedded in large multimodal models to empower the small few-shot learner. Three key challenges arise in achieving this goal: \ding{182} \textit{effectively distilling desired visual and semantic knowledge from these models}, \ding{183} \textit{transforming explicit or implicit knowledge into a usable form}, and \ding{184} \textit{adaptively integrating them with limited visual data to improve FSL performance}. As illustrated in Figure~\ref{fig:overview}, \textsc{SynTrans} addresses these challenges in three stages. First, we introduce a vast CLIP model as a strong teacher, adding a linear projection layer after the frozen few-shot vision encoder as a weak student to distill semantic-aligned visual knowledge via an unsupervised proxy task. Next, our Synergistic Knowledge Mining (SynMine) module exploits a large language model to generate comprehensive text descriptions by tapping into the implicit knowledge through chain-of-thought prompting. These descriptions are then refined by a visual-language model into rich semantic descriptors, producing deeper, context-aware understanding of class characteristics. Central to the semantic transfer stage is the Visual-Semantic Bridging (VSBrid) module,  which leverages a dual encoder-decoder design to facilitate bi-directional knowledge transfer between the visual and semantic spaces, ultimately mapping these high-quality visual embeddings and semantic descriptors to practical class-specific classifier weights. Finally, a visual weight generator and a semantic weight reconstructor are incorporated for dynamic visual-semantic knowledge fusion, constructing robust and adaptable multimodal classifiers.

We evaluate \textsc{SynTrans} on four benchmark datasets, demonstrating its state-of-the-art performance even when equipped with a simple few-shot vision encoder. To the best of our knowledge, \textsc{SynTrans} is the first framework that systematically integrates knowledge from large multimodal models to empower small few-shot learners, opening new avenues for bridging the gap between human-like intuition and machine learning in FSL.

\section{Related Works}
\paragraph{Visual-based FSL Methods.}
Visual-based FSL methods~\cite{0007LYYL023,fu2023styleadv,fu2024cross} transfer prior visual knowledge from base classes to novel classes. Broadly, these approaches can be categorized into two branches: optimization-based methods and metric-based methods. Their core distinction lies in how they leverage the support set, either by fine-tuning an end-to-end network or by directly generating classifiers for novel classes. Optimization-based methods~\cite{finn2017model,ravi2016optimization} focus on learning an effective initialization or optimization strategy, enabling rapid model adaptation to novel tasks with only a few fine-tuning steps. Despite their flexibility, such methods can face meta-overfitting issues when limited labeled data are available. In contrast, metric-based methods~\cite{vinyals2016matching,snell2017prototypical,sung2018learning} learn a metric space where samples from the same category lie close together while those from different categories are farther apart. Another emerging paradigm~\cite{chen2018closer} involves pre-training a powerful feature extractor on base data and directly generating classifier weights for new classes using, for instance, a cosine classifier~\cite{LuoZXWRY18}. However, purely visual approaches may struggle under limited training samples, as real-world data often contain background noise and significant intraclass variation. Consequently, relying solely on visual cues may be insufficient for robust recognition of novel categories. This limitation naturally leads to the question of how high-quality semantic knowledge can be integrated to complement visually dominated FSL methods.

\paragraph{Semantic-based FSL Methods.}
To overcome the shortcomings of purely visual approaches, semantic-based FSL methods~\cite{li2023knowledge,Lu0ZH023} leverage auxiliary semantic information such as attributes~\cite{lampert2009learning}, word embeddings~\cite{XianSSA19}, or even knowledge graphs~\cite{miller1995wordnet}. For example, AM3~\cite{xing2019adaptive} combines label embeddings derived from class names with visual prototypes to generate semantic prototypes, which are then adaptively fused. Knowledge graphs also provide valuable correlation cues: KTN~\cite{PengLZLQT19} builds a graph convolutional network (GCN) with node representations and edges derived from label embeddings and semantic relationships, allowing knowledge transfer from base to novel categories.
Nevertheless, these methods often rely on either sparse attribute annotations or word vectors obtained from limited textual sources (\emph{e.g.}, Glove~\cite{pennington2014glove}, Word2vec~\cite{word2vec}), which may lack sufficient contextual richness. As a result, the potential noise in these external semantics can hinder performance. Encouraged by the rise of large multimodal models~\cite{Ouyang0JAWMZASR22,RadfordKHRGASAM21} that encapsulate abundant implicit knowledge, we investigate how to extract and distill higher-quality prior knowledge to complement visual-based FSL. Our experiments show that the desired knowledge distilled from these large multimodal models can significantly boost FSL performance, even when used with a relatively simple pre-trained backbone.

\section{Method}
\subsection{Problem Formulation}
\label{ProblemFormulation}
We consider the standard FSL setting~\cite{vinyals2016matching}, where two disjoint sets of classes are given: a \emph{base} set $\mathcal{C}_{base}$ and a \emph{novel} set $\mathcal{C}_{novel}$, such that $\mathcal{C}_{base} \cap \mathcal{C}_{novel} = \emptyset$. The model first trains on samples from $\mathcal{C}_{base}$, denoted by $\mathcal{D}_{base} = \{(x, y)\}$, where each pair $(x, y)$ corresponds to an image $x$ and its one-hot label $y$ drawn from $\mathcal{C}_{base}$. For semantic reference, each label $y$ can be mapped to a specific class name $c$, such as ``\texttt{House Finch}'' or ``\texttt{Robin}''. During evaluation, we adopt the ``$N$-way $K$-shot'' protocol, which randomly selects $N$ categories from $\mathcal{C}_{novel}$ to construct two subsets: a support set and a query set. The support set $\mathcal{S} = \{(x_i, y_i)\}_{i=1}^{N \times K}$ contains $K$ examples for each of the $N$ novel classes, while the query set $\mathcal{Q} = \{(x_i, y_i)\}_{i=1}^{N \times Q}$ includes $Q$ samples from the same $N$ categories. Here, $x_i$ denotes the $i$-th image, and $y_i$ is one of the novel class labels in $\mathcal{C}_{novel}$. The main objective of FSL is to leverage the base knowledge (learned from $\mathcal{D}_{base}$) along with the limited support samples in $\mathcal{S}$ to classify new query images in $\mathcal{Q}$.

\subsection{Overall Framework}
As shown in Figure~\ref{fig:overview}, the \textsc{SynTrans} framework consists of three stages: visual knowledge distillation, semantic knowledge transfer, and multi-modal knowledge fusion. These stages work synergistically to enhance a few-shot learner by integrating visual and semantic knowledge. \textbf{Notably, unlike traditional FSL methods, our method does not require fine-tuning the vision encoder at any stages}. This unique characteristic highlights the efficiency of \textsc{SynTrans}, making it readily applicable on top of existing few-shot learners.

\paragraph{Visual Knowledge Distillation.}
In this stage, we train a lightweight projector $f_{\varphi}$ and a cosine classifier $f_{\Phi}$. The projector $f_{\varphi}$ enables \textsc{SynTrans} to perform CLIP-like vision-semantic alignment, while the classifier $f_{\Phi}$ distills more task-relevant visual knowledge from the few-shot vision encoder for subsequent knowledge transfer. Specifically, the parameters of $f_{\Phi}$ are optimized using cross-entropy loss, while the parameters of $f_{\varphi}$ are optimized via the teacher-student distillation paradigm~\cite{wu2024self}.

\paragraph{Semantic Knowledge Transfer.}
This stage consists of two phases: semantic knowledge mining and visual-semantic knowledge bridging. First, we introduce the train-free Synergistic Knowledge Mining (SynMine) module, which efficiently extracts implicit knowledge from visual- and language models to generate high-quality semantic descriptors. Next, the Visual-Semantic Bridging (VSBird) module facilitates bidirectional knowledge transfer between the visual and semantic spaces, mapping the visual embeddings and semantic descriptors to category-specific classifier weights.

\paragraph{Multi-modal Knowledge Fusion.}
In this stage, the frozen few-shot learner computes visual prototypes for all classes in the support set. Simultaneously, the VSBird module generates classifier weights based on these visual prototypes and the semantic descriptors. To integrate these weights into robust multimodal few-shot classifiers, we introduce a visual weight generator and a semantic weight reconstructor, which function as meta-learners, adaptively combining both types of classifier weights for current FSL task.

\subsection{Visual Knowledge Distillation}

In this stage, we introduce a large CLIP teacher model to distill semantic-aligned visual knowledge, empowering the frozen few-shot learner with the ability to perform CLIP-like vision-semantic alignment. Unlike the heavy CLIP vision encoder, the frozen few-shot vision encoder is lightweight and compatible with existing FSL methods, such as IER~\cite{Rizve0KS21} and SMKD~\cite{LinHMH0C23}, enabling training from scratch for simplicity.

As shown in Figure~\ref{fig:overview}(a), we align the few-shot vision encoder with the CLIP vision encoder by learning a linear projector. To achieve this, we treat unsupervised vision-semantic alignment as a proxy task, inspired by~\cite{LiLFZWCY24}, and use knowledge distillation to align the output distributions of both models. For a given FSL task with images $\{x_i\}^{N \times (K+Q)}$ and their class names $\{c_j\}^{N}$ from $\mathcal{C}_{base}$, the CLIP teacher model first processes the category names using a fixed prompt template (e.g., "a photo of a \{\texttt{CLASS}\}"), then passing images and category names through the image encoder $f_I^t$ and text encoder $f_T^t$ to obtain normalized teacher image features $u^t_i$ and text features $w^t_j$. We then input the same images into the frozen few-shot vision encoder $f_I^s$ to obtain the normalized student image features $u^s_i$. The learnable projector $f_{\varphi}(\cdot)$ is introduced to match the feature dimensions at minimal computational cost while ensuring alignment quality. The teacher and student image features, along with the generated teacher text features, are used to compute the output logits $q^t_i$ and $q^s_i$ for the teacher and student models, respectively. The knowledge distillation loss is formulated using Kullback-Leibler divergence:
\begin{equation}\label{eq1}
\mathcal{L}_{kd}\left(q^t, q^s, \tau\right)=\tau^2 K L\left(\sigma\left(q^t / \tau\right), \sigma\left(q^s / \tau\right)\right).
\end{equation}
where $\sigma(\cdot)$ is the softmax function and $\tau$ is the temperature. 

Additionally, the cosine classifier $f_{\Phi}$ is trained to acquire transferable visual knowledge from the base dataset $\mathcal{D}_{base}$ using the pre-trained few-shot vision encoder. Let $W_{base}=\{\mathbf{w}^{c}\}_{c\in\mathcal{C}_{base}}$ denotes the weight vectors of classifier $f_{\Phi}$.
The classification loss, $\mathcal{L}_{ce}$, is computed with cross-entropy over the base classes $\mathcal{C}_{base}$:
\begin{equation}\label{eq2}
\mathcal{L}_{ce}=-\log\frac{\exp \left( \cos (f_I^s(x_i), \mathbf{w}^{c}) \right)}{\sum_{c^{\prime}=1}^{|\mathcal{C}_{base}|} \exp \left( \cos f_I^s(x_i), \mathbf{w}^{c^{\prime}} \right)}.
\end{equation}
This allows the learned classifier can provide more meaningful and task-relevant visual knowledge for subsequent knowledge transfer and fusion within our \textsc{SynTrans}. 
Finally, we combine the knowledge distillation and classification losses into a multi-task objective to ensure the projector $f_{\varphi}$ and classifier $f_{\Phi}$ perform their respective tasks effectively within the same FSL task as $\mathcal{L}_{vis} = \mathcal{L}_{ce} + \mathcal{L}_{kd}$.

\subsection{Synergistic Knowledge Mining}
\label{sec:SynMine}

In leveraging the capabilities of large multimodal models, our proposed SynMine first utilizes the rich common-sense knowledge embedded in large language model (LLM) to generate detailed class descriptions. Additionally, SynMine takes advantage of the advanced image-text alignment capabilities of pre-trained visual-language model (VLM) to refine these descriptions into high-quality semantic descriptors, enhancing their relevance for FSL.

\begin{figure}[t]
\begin{center}
\includegraphics[width=0.4\textwidth]{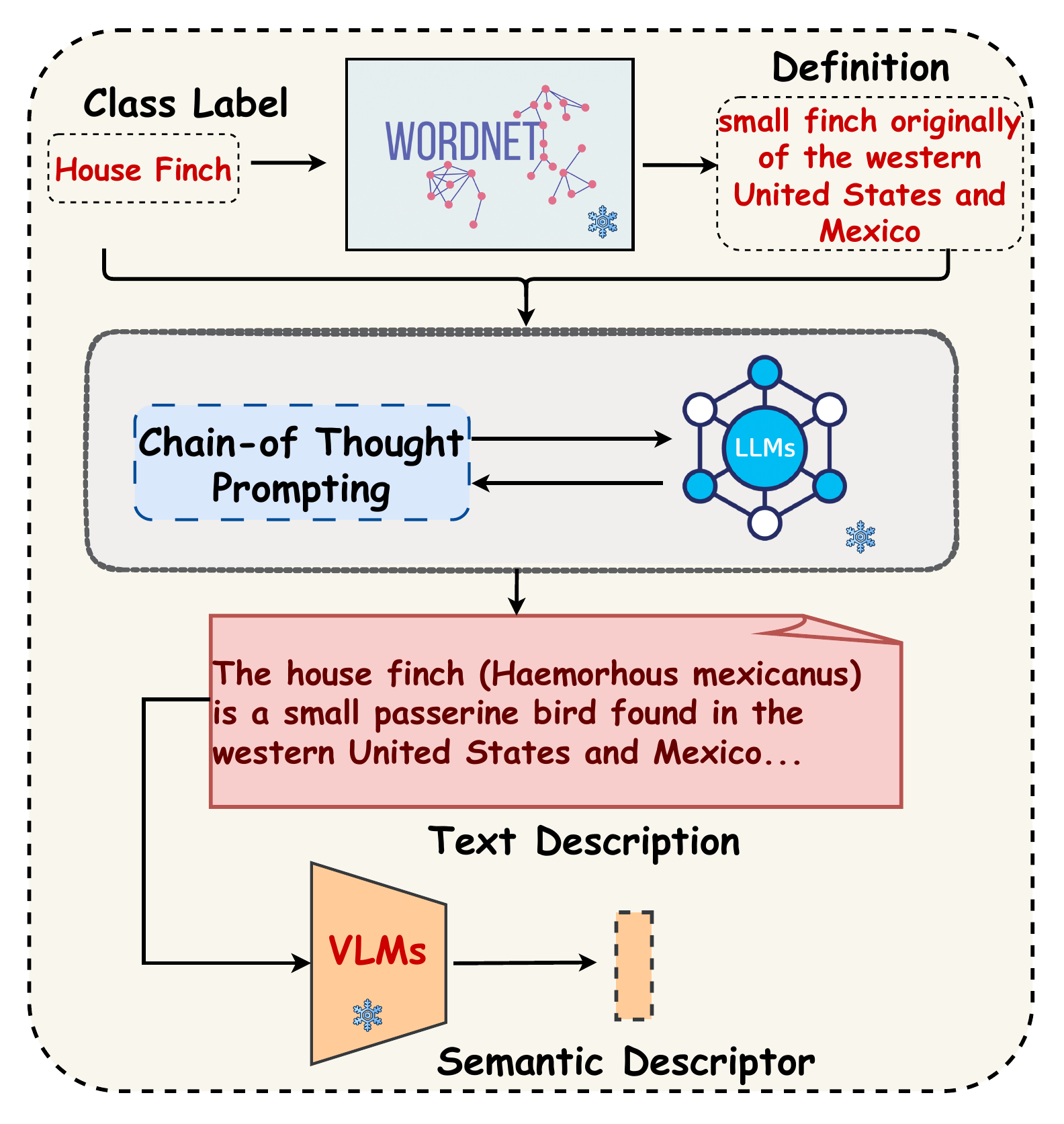}
\end{center}
  \caption{
   The pipeline of how the proposed SynMine module generates high-quality semantic descriptors.
  }
\label{fig:mckm}
\end{figure}

As illustrated in Figure~\ref{fig:mckm}, SynMine follows a multi-step process to enhance the comprehensive understanding of class characteristics. Inspired by the ``chain-of-thought'' prompting technique~\cite{Wei0SBIXCLZ22}, which improves LLM performance through intermediate reasoning, we guide LLM using this technique. Initially, a concise definition matching the visual content is provided to the LLM to eliminate ambiguity and help it generate class descriptions focused on visual features. The first prompt is as follows:

\begin{tcolorbox}[width=1.0\linewidth, sharp corners, center title]
{\bf Prompt 1:} [\texttt{DEFINITION}] is the definition of the [\texttt{CLASS}]. Can you describe the visual features associated with this category?
\end{tcolorbox} 

Here, [\texttt{DEFINITION}] is a brief class definition obtained from WordNet~\cite{miller1995wordnet}, and [\texttt{CLASS}] refers to the class name. Next, we refine these descriptions by focusing on distinctive visual attributes based on the initial responses. The second prompt is as follows:

\begin{tcolorbox}[width=1.0\linewidth, sharp corners, center title]
{\bf Prompt 2:} Please describe the [\texttt{CLASS}] in a maximum of five sentences, focusing on discriminative visual features. Make the description more detailed and aligned with scientific facts, avoiding general summaries and subjective interpretations.
\end{tcolorbox}

The generated descriptions are then processed by the text encoder of a pre-trained VLM, producing high-quality semantic descriptors. This approach is preferred over traditional word embedding models, such as Word2Vec~\cite{word2vec}, as it captures deeper contextual meanings and nuances, enhancing the discrimination of visual features across classes.

\subsection{Bidirectional Visual-Semantic Bridging}

\begin{figure}[t]
\begin{center}
\includegraphics[width=0.48\textwidth]{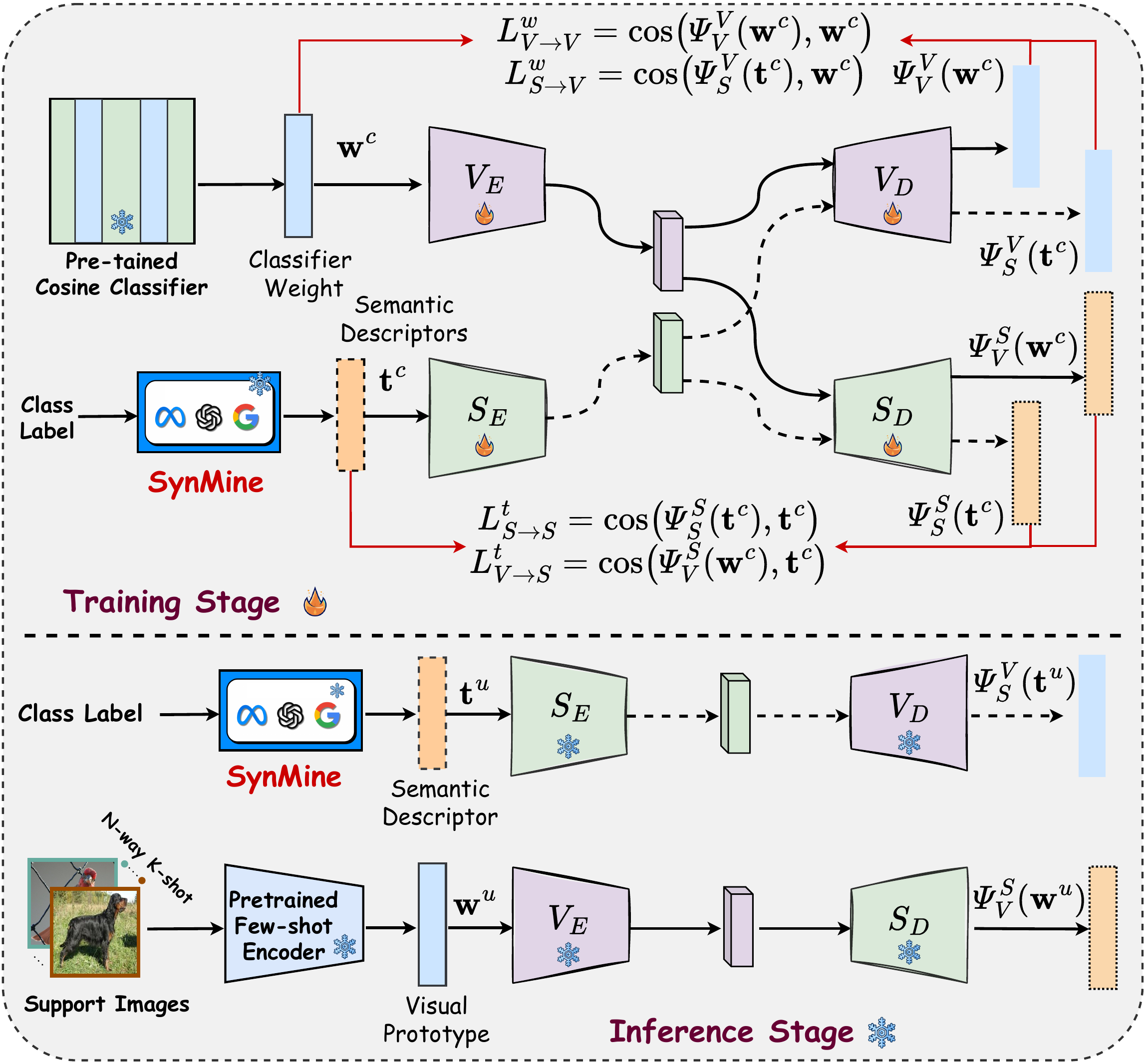}
\end{center}
  \caption{
   The pipeline of the proposed \textbf{V}isual-\textbf{S}emantic \textbf{Brid}ging (VSBird) module.
  }
\label{fig:VSKB}
\end{figure}

After distilling high-quality semantic descriptors through the SynMine module, we introduce the Visual-Semantic Bridging (VSBird) module, as shown in Figure~\ref{fig:VSKB}. VSBird employs a dual autoencoder architecture to establish bidirectional mappings within and between the visual and semantic spaces, aiming to reconstruct multimodal classifier weights based on visual embeddings and semantic descriptors. This step is crucial for mapping visual and semantic knowledge into actionable classifier weights for FSL.

In this module, the classifier weights $f_{\Phi}$ are treated as visual embeddings encoding distinctive visual knowledge. We then use the corresponding semantic descriptors to facilitate the visual-semantic bridging. Specifically, let $W_{base}=\{\mathbf{w}^{c}\}_{c\in\mathcal{C}_{base}}$ denote the classifier weights and $T_{base}=\{\mathbf{t}^{c}\}_{c\in\mathcal{C}_{base}}$ the semantic descriptors. Our goal is to learn a mapping $\Psi_{S}^{V}:\mathcal{S} \rightarrow \mathcal{V}$, where $\mathcal{S}$ and $\mathcal{V}$ represent the semantic and visual spaces, respectively. The VSBird architecture consists of two encoder-decoder subnetworks: the visual encoder $V_{E}: \mathcal{V} \rightarrow \mathcal{Z}$, the semantic encoder $S_{E}: \mathcal{S} \rightarrow \mathcal{Z}$, the visual decoder $V_{D}: \mathcal{Z} \rightarrow \mathcal{V}$, and the semantic decoder $S_{D}: \mathcal{Z} \rightarrow \mathcal{S}$, with $\mathcal{Z}$ as the latent space. These components define the desired semantics-to-weights mapping as $\Psi_{S}^{V}(\mathbf{t}^{c}) = V_{D}(S_{E}(\mathbf{t}^{c}))$. To avoid bias towards base classes, we introduce self- and cross-reconstruction objectives. These ensure both modalities preserve their structures within their respective spaces while also aligning the latent spaces. The self-reconstruction objectives minimize the following terms:
\begin{equation}\label{eq6}
\resizebox{0.91\hsize}{!}{$
\begin{aligned}
\mathcal{L} _{V \rightarrow V}^w &= \cos\left(\Psi_{V}^{V}\left(\mathbf{w}^c\right), \mathbf{w}^c\right)= \cos\left(V_{D}\left(V_{E}(\mathbf{w}^c)\right), \mathbf{w}^c\right),\\
\mathcal{L} _{S \rightarrow S}^t &= \cos\left(\Psi_{S}^{S}\left(\mathbf{t}^c\right), \mathbf{t}^c\right)=\cos\left(S_{D}\left(S_{E}(\mathbf{t}^c)\right), \mathbf{t}^c\right).
\end{aligned}$}
\end{equation}
Here, $\cos(\cdot,\cdot)$ denotes the cosine distance function. While the two autoencoders preserve structure within their respective spaces, they do not ensure alignment between the two latent spaces. To address this, we introduce two additional cross-reconstruction objectives in a symmetric manner:
\begin{equation}\label{eq7}
\begin{aligned}
\mathcal{L} _{S \rightarrow V}^w &= \cos\left(\Psi_{S}^{V}\left(\mathbf{t}^c\right), \mathbf{w}^c\right)= \cos\left(V_{D}\left(S_{E}(\mathbf{t}^c)\right), \mathbf{w}^c\right),\\
\mathcal{L} _{V \rightarrow S}^t &= \cos\left(\Psi_{V}^{S}\left(\mathbf{w}^c\right), \mathbf{t}^c\right)=\cos\left(S_{D}\left(V_{E}(\mathbf{w}^c)\right), \mathbf{t}^c\right).
\end{aligned}
\end{equation}

The final objective of VSBird is a balanced combination of these reconstruction terms:
\begin{equation}\label{eq8}
\mathcal{L}_{t}^{w} = \alpha*(\mathcal{L} _{V \rightarrow V}^w + \mathcal{L} _{S \rightarrow S}^t) + (1-\alpha)*(\mathcal{L} _{S \rightarrow V}^w + \mathcal{L} _{V \rightarrow S}^t),
\end{equation}
where $\alpha$ is a weight coefficient used to control the balance between self-reconstruction and cross-reconstruction. 

During inference, for a novel class $u \in \mathcal{C}_{novel}$,
the semantic encoder $S_{E}$ and visual decoder $V_{D}$ infer the semantic-derived classifier weight:
\begin{equation}\label{eq9}
\mathbf{w}^{u}_{s^{'}}=\Psi_{S}^{V}(\mathbf{t}^{u})=V_{D}(S_{E}(\mathbf{t}^{u})),
\end{equation}
and for a unseen visual prototype $\mathbf{w}^u$, the visual encoder $V_{E}$ and semantic decoder $S_{D}$ infer the visual-derived classifier weight:
\begin{equation}\label{eq10}
\mathbf{w}^{u}_{v^{'}}=\Psi_{V}^{S}(\mathbf{w}^{u})=S_{D}(V_{E}(\mathbf{w}^{u})).
\end{equation}

\subsection{Multi-modal Knowledge Fusion}
In the knowledge fusion stage, we create $N$-way $K$-shot meta-tasks from the base training set $\mathcal{D}_{base}$ to mimic the few-shot scenario during testing. The main goal is to develop a visual weight generator and a semantic weight reconstructor to combine visual-based and semantic-based classifier weights, forming robust multimodal classifiers for FSL tasks.

Given the pre-trained few-shot vision encoder $f_I^s$, we calculate the visual-based classifier weight $\mathbf{w}^{m}_{v}$ for each class $m$ in the support set $\mathcal{S}$ as:
$\mathbf{w}^{m}_{v} = \frac{1}{\left\| \sum_{i=1}^{K} f_I^s(x_{i}) \right\|_{2}} \sum_{i=1}^{K} f_I^s(x_{i})$, where $K$ is the number of samples per class $m$. Next, similar to Equations~\eqref{eq9} and \eqref{eq10}, we transfer visual and semantic knowledge from SynMine to generate visual-derived classifier weight $\mathbf{w}^{m}_{v^{'}}$ and semantic-derived classifier weight $\mathbf{w}^{m}_{s^{'}}$ for each class $m \in \mathcal{S}$. These weights complement each other: the visual-based weight $\mathbf{w}^{m}_{v}$ and semantic-derived classifier weight $\mathbf{w}^{m}_{s^{'}}$ complement one another, and similarly, the semantic-based weight $\mathbf{w}^{m}_{s}$ and visual-derived weight $\mathbf{w}^{m}_{v^{'}}$ complement each other.

To facilitate visual-semantic weight fusion, we introduce the visual weight generator $G$ and semantic weight reconstructor $R$. The generator $G$, consisting of a fully connected layer followed by a sigmoid function, adaptively produces a weight coefficient as $\beta = \frac{1}{1 + \exp \left(-G\left(\mathbf{w}^{m}_{s^{'}}\right)\right)}$, with values restricted to the range $[0,1]$. This coefficient $\beta$ is used to balance the contributions of $\mathbf{w}^{m}_{s^{'}}$ and $\mathbf{w}^{m}_{v}$ in the visual-dominated classifier as $\mathbf{w}^{m}_{V} = \beta \cdot \mathbf{w}^{m}_{s^{'}} + (1-\beta) \cdot \mathbf{w}^{m}_{v}$. Similarly, the semantic-based weight $\mathbf{w}^{m}_{s}$ and visual-derived weight $\mathbf{w}^{m}_{v^{'}}$ are concatenated and passed through the reconstructor $R$ to form the semantic-dominated classifier as $\mathbf{w}^{m}_{S} = R\left(\mathbf{w}^{m}_{s}, \mathbf{w}^{m}_{v^{'}}\right) = \sigma\left(\left[\mathbf{w}^{m}_{s} \cdot \mathbf{w}^{m}_{v^{'}}\right]^{\top} W_1\right) W_2$, where $\sigma$ is an activation function, and $W_1$ and $W_2$ are learnable weights. Afterward, we generate multimodal classifiers for the $N$-way $K$-shot meta-task: $\mathbf{W}_{V}=\{\mathbf{w}^{m}_{V}\}_{m=1}^{N}$ and $\mathbf{W}_{S}=\{\mathbf{w}^{m}_{S}\}_{m=1}^{N}$. Given an image $q$ from the query set, we compute the probabilities $P_v$ and $P_s$ using cosine similarity as $P_v = \frac{\exp \left( \cos(f_I^s(q), \mathbf{w}^{m}_{V}) \right)}{\sum_{m^{\prime}=1}^{N} \exp \left( \cos(f_I^s(q), \mathbf{w}_{V}^{m^{\prime}}) \right)}$ and $P_s = \frac{\exp \left( \cos(f_{\varphi}(f_I^s(q)), \mathbf{w}^{m}_{S}) \right)}{\sum_{m^{\prime}=1}^{N} \exp \left( \cos(f_{\varphi}(f_I^s(q)), \mathbf{w}_{S}^{m^{\prime}}) \right)}$. We use the softmax cross-entropy loss function to train the generator $G$ and reconstructor $R$.

\paragraph{Inference.}
During inference, the visual-dominated classifier and the
semantic-dominated classifier are complementary to each other.
Therefore, we propose a fusion mechanism to obtain the final class logits. Given a test image $x_t$, the
prediction is made as follows:
\begin{equation} 
y^* = \arg \max \left( \left\langle \left[ f_I^s(x_t), f_{\varphi}(f_I^s(x_t)) \right], \left[ \mathbf{W}_V, \lambda \mathbf{W}_S \right] \right\rangle \right),
\end{equation}
where $\lambda$ is a positive balancing coefficient, empirically set to $\frac{1}{K}$ in our \textsc{SynTrans}.

\section{Experiments}
\subsection{Datasets}
Following the settings in~\cite{Zhang0J024},
we evaluate the performance of the proposed \textsc{SynTrans} framework on four widely used benchmarks in FSL. Two of these datasets are derived from the ImageNet dataset~\cite{russakovsky2015imagenet}: MiniImageNet~\cite{vinyals2016matching} and TieredImageNet~\cite{RenTRSSTLZ18}. Another two dataset are CIFAR-FS~\cite{lee2019meta} and FC100~\cite{oreshkin2018tadam}.

\begin{table}[t!]
  \centering
  \resizebox{\linewidth}{!}{%
		\begin{tabular}{c l c c c c c}
    \toprule
        &\multirow{2}{*}{\textbf{Method}} 
        & \multirow{2}{*}{\textbf{Backbone}} 
        & \multicolumn{2}{c}{\textbf{MiniImageNet}} 
        & \multicolumn{2}{c}{\textbf{TieredImageNet}}\\ 
         \cmidrule(lr){4-5} \cmidrule(lr){6-7}
        & & & \textbf{5-way 1-shot} 
        & \textbf{5-way 5-shot}  
        & \textbf{5-way 1-shot} 
        & \textbf{5-way 5-shot}\\ 
        \midrule
        \multirow{9}{*}{\rotatebox{90}{\textbf{Visual-Based}}}
        &MatchNet \cite{vinyals2016matching} & ResNet-12 & 65.64 ± {\footnotesize 0.20} & 78.72 ± {\footnotesize 0.15} & 68.50 ± {\footnotesize 0.92} & 80.60 ± {\footnotesize 0.71} \\
        &ProtoNet \cite{snell2017prototypical} & ResNet-12 & 62.29 ± {\footnotesize 0.33} & 79.46 ± {\footnotesize 0.48} & 68.25 ± {\footnotesize 0.23} & 84.01 ± {\footnotesize 0.56} \\
        &MAML \cite{finn2017model} & ResNet-12  & 58.05 ± {\footnotesize 0.21} & 58.05 ± {\footnotesize 0.10} & 67.92 ± {\footnotesize 0.17} & 72.41 ± {\footnotesize 0.20} \\
        &MetaOptNet \cite{lee2019meta} & ResNet-18 & 62.64 ± {\footnotesize 0.61} & 78.63 ± {\footnotesize 0.46} & 65.99 ± {\footnotesize 0.72} & 81.56 ± {\footnotesize 0.53} \\
        &FEAT \cite{ye2020few}  & ResNet-12 & 66.78 ± {\footnotesize 0.20} & 82.05 ± {\footnotesize 0.14} & 70.80 ± {\footnotesize 0.23} & 84.79 ± {\footnotesize 0.16}\\
        &Meta-Baseline \cite{Chen_2021_ICCV}  & ResNet-12 & 63.17 ± {\footnotesize 0.23} & 79.26 ± {\footnotesize 0.17} & 68.62 ± {\footnotesize 0.27} & 83.29 ± {\footnotesize 0.18} \\
        & CVET \cite{YangWZ22} & ResNet-12 & 70.19 ± {\footnotesize 0.46} & 84.66 ± {\footnotesize 0.29} & 72.62 ± {\footnotesize 0.51} & 86.62 ± {\footnotesize 0.33} \\
        &FGFL \cite{cheng2023frequency}  & ResNet-12 & 69.14 ± {\footnotesize 0.80} & 86.01 ± {\footnotesize 0.62} & 73.21 ± {\footnotesize 0.88} & 87.21 ± {\footnotesize 0.61}\\
        & SUN~\cite{DongZYZ22} & ViT-S & 67.80 ± {\footnotesize 0.45} & 83.25 ± {\footnotesize 0.30} & 72.99 ± {\footnotesize 0.50} & 86.74 ± {\footnotesize 0.33}\\
        & SMKD~\cite{LinHMH0C23} & ViT-S & 74.28 ± {\footnotesize 0.18} & 88.82 ± {\footnotesize 0.09} & 78.83 ± {\footnotesize 0.20} & 91.02 ± {\footnotesize 0.12}\\
        & FewTURE~\cite{HillerMHD22} & Swin-T & 72.40 ± {\footnotesize 0.78} & 86.38 ± {\footnotesize 0.49} & 76.32 ± {\footnotesize 0.87} & 89.96 ± {\footnotesize 0.55}\\
        \midrule
        \multirow{8}{*}{\rotatebox{90}{\textbf{Semantic-Based}}}
        &KTN \cite{PengLZLQT19}  & ResNet-12 & 61.42 ± {\footnotesize 0.72} & 70.19 ± {\footnotesize 0.62} & 68.01 ± {\footnotesize 0.73} & 79.06 ± {\footnotesize 0.70} \\
        &AM3 \cite{xing2019adaptive}  & ResNet-12 & 65.30 ± {\footnotesize 0.49} & 78.10 ± {\footnotesize 0.36} & 69.08 ± {\footnotesize 0.47} & 82.58 ± {\footnotesize 0.31} \\
        &PC-FSL \cite{ZhangLYHZ21}  & ResNet-12 & 69.68 ± {\footnotesize 0.76} & 81.65 ± {\footnotesize 0.54} & 74.19 ± {\footnotesize 0.90}& 86.09 ± {\footnotesize 0.60} \\
        &SEGA \cite{YangWC21}  & ResNet-12 & 69.04 ± {\footnotesize 0.26} & 79.03 ± {\footnotesize 0.18} & 72.18 ± {\footnotesize 0.30} & 84.28 ± {\footnotesize 0.21}\\
        &LPE-CLIP \cite{YangWC23}  & ResNet-12 & 71.64 ± {\footnotesize 0.40} & 79.67 ± {\footnotesize 0.32} & 73.88 ± {\footnotesize 0.48} & 84.88 ± {\footnotesize 0.36}\\
        &\textsc{KSTNet} \cite{li2023knowledge} & ResNet-12 & 71.51 ± {\footnotesize 0.73} & 82.61 ± {\footnotesize 0.48} & 75.52 ± {\footnotesize 0.77} & 85.85 ± {\footnotesize 0.59} \\ 
        &4S-FSL \cite{Lu0ZH023}  & ResNet-12 & 72.64 ± {\footnotesize 0.70} & 84.73 ± {\footnotesize 0.50} & - & - \\ 
        &SP-CLIP~\cite{ChenS0WWT23}  & Visformer-T & 72.31 ± {\footnotesize 0.40} & 83.42 ± {\footnotesize 0.30} & 78.03 ± {\footnotesize 0.46} & 88.55 ± {\footnotesize 0.32} \\ 
        &SemFew~\cite{Zhang0J024}  & Swin-T & 78.94 ± {\footnotesize 0.66} & 86.49 ± {\footnotesize 0.50} & 82.37 ± {\footnotesize 0.77} & 89.89 ± {\footnotesize 0.52} \\ 
        \cmidrule(lr){2-7}
        &\textsc{SynTrans} & ResNet-12 & 76.20 ± {\footnotesize 0.69} & 86.12 ± {\footnotesize 0.54} & { 79.69 ± {\footnotesize 0.81}} & {  87.78 ± {\footnotesize 0.60 }} \\
        &\textsc{SynTrans} & ViT-S & \textbf{ 81.30 ± {\footnotesize 0.61}} & \textbf{ 89.96 ± {\footnotesize 0.42}} & \textbf{  84.31 ± {\footnotesize 0.54}} & \textbf{ 91.73 ± {\footnotesize 0.44}} \\
        \bottomrule
    \end{tabular}
	}
    \caption{Comparison with state-of-the-art methods on MiniImageNet and TieredImageNet.
  }
  \label{tab:1}
\end{table}

\subsection{Implementation Details}
\paragraph{Architecture.} 
In all experimental setups, we utilize ResNet-12 and ViT-Small (ViT-S) as the few-shot vision encoders. Specifically, the ResNet-12 encoder is pretrained using the training strategy described in IER~\cite{Rizve0KS21}, while the ViT-S encoder follows the strategy reported in SMKD~\cite{LinHMH0C23}. For the ResNet-12 encoder, visual features are obtained by averaging the outputs from the final residual block, resulting in a feature dimension of $640$. For the ViT-S encoder, visual embeddings are computed by averaging the hidden states from the last transformer block, yielding a feature dimension of $384$. During the visual knowledge distillation stage, we use both the vision and text encoders from Res50x4 CLIP~\cite{RadfordKHRGASAM21} as a strong teacher and employ two linear layers to construct the linear projector $f_{\varphi}$. In the SynMine module, we leverage the GPT-3.5-turbo model as the large language model and Res50x4 CLIP~\cite{RadfordKHRGASAM21} as the visual-language model. The SynMine module facilitates the alignment of features across vision and language modalities. The encoders and decoders of the VSBird module consist of single-layer linear mappings, with ReLU activation following the encoder mappings. A simple fully connected layer serves as the learnable weight generator $G$. The weight reconstructor $R$ combines visual and textual features using two fully connected layers followed by a LeakyReLU activation function. The hidden layer has a dimension of $2048$.

\paragraph{Training Details.}
During the visual knowledge distillation stage, we freeze both the few-shot vision encoder and CLIP’s vision encoder, focusing on optimizing the linear projector and cosine classifier. For ResNet-12, we follow the methods in~\cite{li2023knowledge} and resize the input images to $84 \times 84$. For ViT-S, we resize the input image to $320 \times 320$ for MiniImageNet and TieredImageNet, and to $224 \times 224$ for CIFAR-FS and FC100, maintaining consistency with SMKD~\cite{LinHMH0C23}. In the knowledge transfer stage, we freeze all parameters in the SynMine module and only train the parameters of the VSBird module. In the meta-training stage, we train only the parameters of the weight generator and weight reconstructor. For both stages, we employ the Adam optimizer~\cite{KingmaB14adam} with an initial learning rate of $0.0001$ and weight decay of $5 \times 10^{-4}$. In particular, the VSBird module is trained for $50$ epochs with the hyperparameter $\alpha$ set to $0.7$, while the weight generator and weight reconstructor are trained for $10$ epochs. 

\paragraph{Evaluation protocol.} 
The proposed method is evaluated under $5$-way $1$/$5$-shot settings on the novel dataset, with $600$ few-shot tasks randomly sampled from it. Each task consists of 15 query samples per class. We report the average accuracy (\%) with $95\%$ confidence intervals.

\begin{table}[t!]
	\centering
	\resizebox{\linewidth}{!}{%
		\begin{tabular}{l c c c c c}
    \toprule
        \multirow{2}{*}{\textbf{Method}} & \multirow{2}{*}{\textbf{Backbone}} & \multicolumn{2}{c}{\textbf{CIFAR-FS}} & \multicolumn{2}{c}{\textbf{FC100}}\\ 
        \cmidrule(lr){3-4} \cmidrule(lr){5-6}
        & & \textbf{5-way 1-shot} & \textbf{5-way 5-shot} & \textbf{5-way 1-shot} & \textbf{5-way 5-shot}\\ 
        \midrule
        ProtoNet~\cite{snell2017prototypical} & ResNet-12 & 72.20 ± {\footnotesize 0.73} & 83.50 ± {\footnotesize 0.50} & 41.54 ± {\footnotesize 0.76} & 57.08 ± {\footnotesize 0.76}\\
        MetaOptNet~\cite{lee2019meta} & ResNet-12 & 72.80 ± {\footnotesize 0.70} & 84.30 ± {\footnotesize 0.50} & 47.20 ± {\footnotesize 0.60} & 55.50 ± {\footnotesize 0.60}\\
        RFS \cite{tian2020rethinking} & ResNet-12 & 71.50 ± {\footnotesize 0.80} & 86.90 ± {\footnotesize 0.50} & 42.60 ± {\footnotesize 0.70} & 59.10 ± {\footnotesize 0.60}\\
        SUN \cite{DongZYZ22} & ViT-S & 78.37 ± {\footnotesize 0.46} & 88.84 ± {\footnotesize 0.32} & - & -\\
        SMKD \cite{LinHMH0C23} & ViT-S & 80.08 ± {\footnotesize 0.18} & 90.63 ± {\footnotesize 0.13} & 50.38 ± {\footnotesize 0.16} & 68.37 ± {\footnotesize 0.16}\\
        FewTURE \cite{HillerMHD22} &Swin-T & 77.76 ± {\footnotesize 0.81} & 88.90 ± {\footnotesize 0.59} & 47.68 ± {\footnotesize 0.78} & 63.81 ± {\footnotesize 0.75} \\
        \midrule
        SEGA \cite{YangWC21} & ResNet-12 & 78.45 ± {\footnotesize 0.24} & 86.00 ± {\footnotesize 0.20} & - & -\\
        LPE-CLIP \cite{YangWC21} & ResNet-12 & 80.62 ± {\footnotesize 0.41} & 86.22 ± {\footnotesize 0.33} & - & -\\
        4S-FSL \cite{Lu0ZH023} & ResNet-12 & 74.50 ± {\footnotesize 0.84} & 88.76 ± {\footnotesize 0.53}& - & -\\ 
        SP-CLIP \cite{ChenS0WWT23} & Visformer-T & 82.18 ± {\footnotesize 0.40} & 88.24 ± {\footnotesize 0.32}& 48.53 ± {\footnotesize 0.38} & 61.55 ± {\footnotesize 0.41}\\ 
        SemFew \cite{Zhang0J024} & Swin-T & 84.34 ± {\footnotesize 0.67} & 89.11 ± {\footnotesize 0.54}& 54.27 ± {\footnotesize 0.77} & 65.02 ± {\footnotesize 0.72}\\ 
        \cmidrule(lr){1-6}
        \textsc{SynTrans} & ResNet-12 & {82.58 ± {\footnotesize 0.75}} & {89.42 ± {\footnotesize 0.56}} & {52.30 ± {\footnotesize 0.75}} & {64.91 ± {\footnotesize 0.59}} \\
        \textsc{SynTrans} & ViT-S & \textbf{84.64 ± {\footnotesize 0.65}} & \textbf{90.81± {\footnotesize 0.41}}  & \textbf{56.38 ± {\footnotesize 0.69}} & \textbf{69.45± {\footnotesize 0.54}} \\
        \bottomrule
    \end{tabular}
	}
     \caption{Comparison with state-of-the-art methods on CIFAR-FS and FC100.}
     \label{tab:2}
\end{table}

\subsection{Benchmark Comparisons and Evaluations}
Tables~\ref{tab:1} and~\ref{tab:2} summarize the performance of recent state-of-the-art FSL methods on the MiniImageNet, TieredImageNet, CIFAR-FS, and FC100 datasets, focusing on the 5-way 1/5-shot tasks. The experimental results demonstrate that the \textsc{SynTrans} framework achieves outstanding performance across all datasets, particularly when visual information is limited. In the 5-way 1-shot scenario, \textsc{SynTrans} outperforms the most relevant semantic-based method, SenFew~\cite{Zhang0J024}, by a margin of $2.98\%$ due to its flexible knowledge transfer framework that mines rich knowledge from diverse large models. Notably, \textsc{SynTrans} shows greater improvement in the 1-shot setting compared to the 5-shot setting. In the 5-way 5-shot scenario, \textsc{SynTrans} still maintains an advantage over all state-of-the-art methods, though the improvements are less pronounced than in the 1-shot setting. This suggests that even with more visual data, high-quality semantic knowledge from large models can still enhance performance. Overall, the results highlight the effectiveness of \textsc{SynTrans} in various scenarios, demonstrating its ability to leverage both semantic and visual knowledge.

\subsection{Ablation Studies}

\begin{table}[t!]
	\centering
	\resizebox{\linewidth}{!}{%
		\begin{tabular}{l c c c }
    \toprule
        \multirow{2}{*}{\textbf{Knowledge Source}} & \multirow{2}{*}{\textbf{Knowledge Encoder}} & \multicolumn{2}{c}{\textbf{ResNet-12}}\\
        & & \textbf{5-way 1-shot} & \textbf{5-way 5-shot} \\
        \midrule
        Class Names & Word2vec & 72.66 ± {0.70} & 84.68 ± {0.50} \\
        Class Names & VLMs & 72.87 ± {0.71} & 84.67 ± {0.50}\\
        Short Definitions (WordNet) & Word2vec & 72.03 ± {0.73} & 84.18 ± {0.52} \\
        Short Definitions (WordNet) & VLM & 73.52 ± {0.73} & 85.26 ± {0.51} \\
        Rich Descriptions (LLM) & Word2vec & 74.41 ± {0.70} & 85.76 ± {0.51} \\
        Rich Descriptions (LLM) & VLM & \bf 76.20 ± {\footnotesize 0.69} & \bf 86.12 ± {\footnotesize 0.54} \\
        \bottomrule
    \end{tabular}
	}
    \caption{Ablation study about knowledge quality on MiniImageNet.}
	\label{tab:4}
\end{table}

\paragraph{Influence of Semantic Knowledge Quality.} 
As shown in Table~\ref{tab:4}, we evaluate the impact of knowledge quality using different sources and encoders.
First, we compare common semantics (class names) with generic descriptions from WordNet (\emph{i.e.,~}``Short Definitions'') and richer descriptions from LLMs via SynMine (\emph{i.e.,~}``Rich Descriptions''). Results show that LLM-generated descriptions yield the best performance. This suggests that rich descriptions provide deeper semantic understanding, adding nuanced attributes to the classifier weights that simple class names cannot. Next, we compare Word2Vec models with VLMs for encoding semantic knowledge. VLMs outperform Word2Vec in both 1-shot and 5-shot settings, benefiting from their multi-modal training, which enhances semantic understanding. The strong performance of VLMs, particularly when paired with LLM-generated descriptions, highlights the effectiveness of combining large models for optimal knowledge transfer.

\begin{figure}[t!]
    \centering
    \subfloat[5-way 1-shot]{
		\begin{minipage}{0.47\linewidth}
			\centering
			\includegraphics[width=\textwidth]{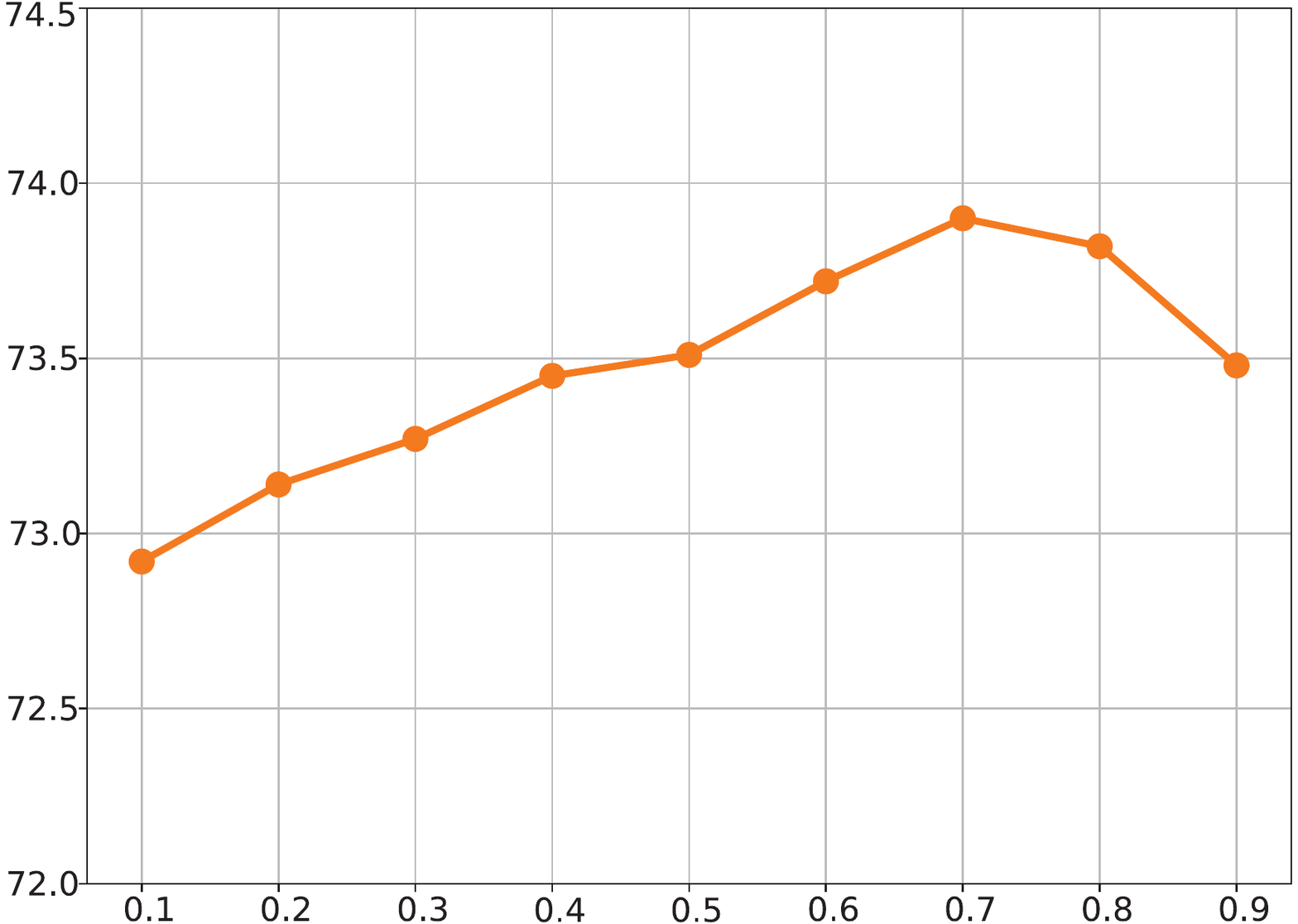}
		\end{minipage}
	}
	\subfloat[5-way 5-shot]{
		\begin{minipage}{0.47\linewidth}
			\centering
			\includegraphics[width=\textwidth]{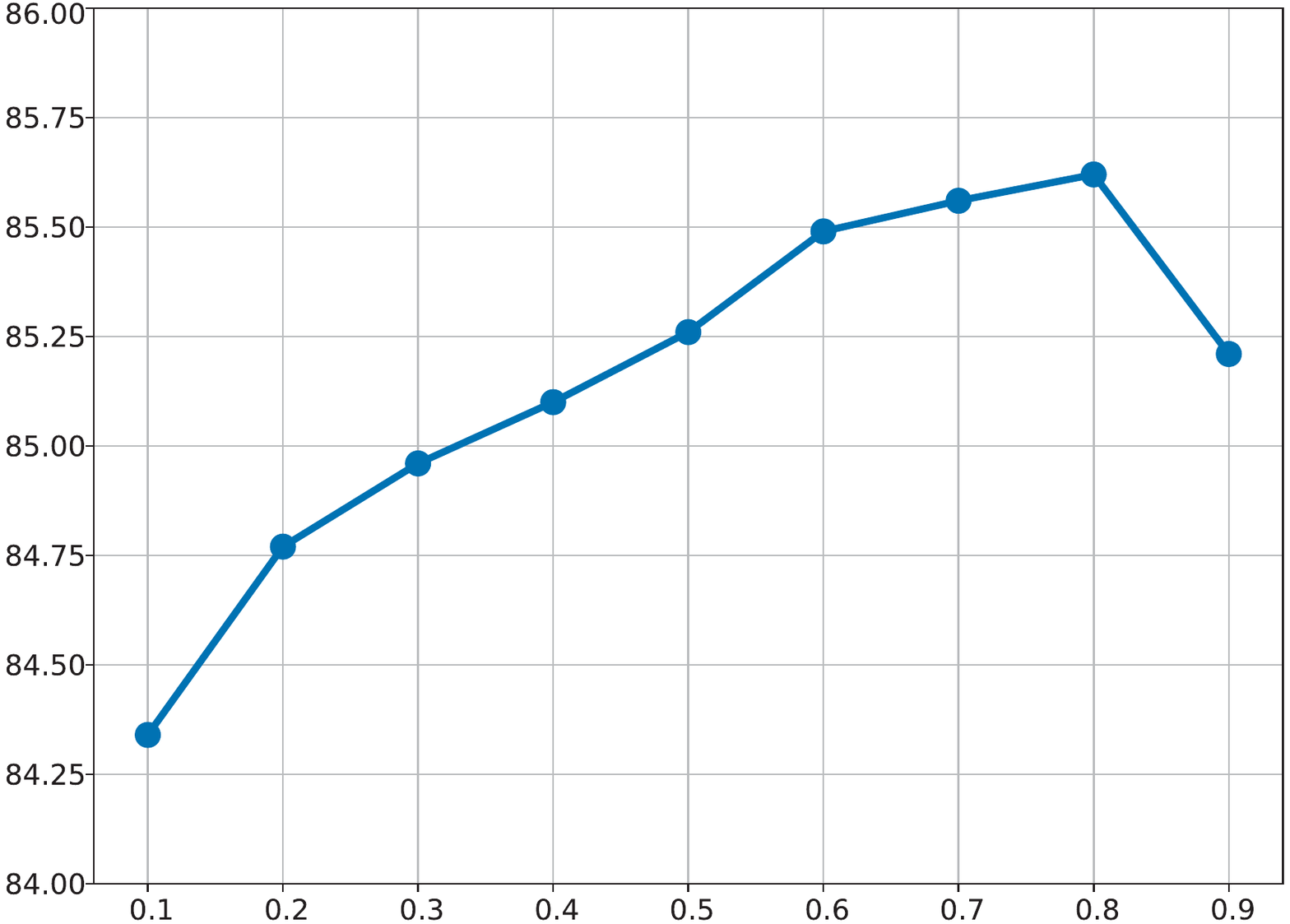}
		\end{minipage}
	}
    \caption{Influence of weight coefficient $\alpha$ on MiniImageNet.}
    \label{fig:shots}
\end{figure}

\paragraph{Influence of Hyper-parameter $\alpha$.}
The VSBird module is essential for bridging visual and semantic spaces, where the dual autoencoder architecture encourages both self-reconstruction within each space and cross-reconstruction between spaces. Figure~\ref{fig:shots} shows the performance of the visual-dominated classifier $\mathbf{W}_{V}$ on MiniImageNet with varying weight coefficients $\alpha$ in Equation~\eqref{eq8}. A larger $\alpha$ enhances the significance of self-reconstruction loss during VSKB module training. When $\alpha$ is too small, self-reconstruction loss is suppressed by cross-reconstruction loss, leading to lower accuracy. This suggests that while cross-reconstruction is crucial for aligning latent spaces, preserving the individual structures of visual and semantic spaces is equally important. However, beyond $\alpha = 0.7$ for the 1-shot setting and $\alpha = 0.8$ for the 5-shot setting, accuracy slightly decreases, indicating that excessive self-reconstruction diminishes the benefits of cross-reconstruction. Thus, balancing self-reconstruction and cross-reconstruction is crucial for optimal performance, with $\alpha = 0.7$ achieving the best trade-off across settings.

\begin{figure}[t!]
	\centering
	\subfloat[5-way 1-shot]{
		\begin{minipage}{0.45\linewidth}
			\centering
			\includegraphics[width=0.9\textwidth]{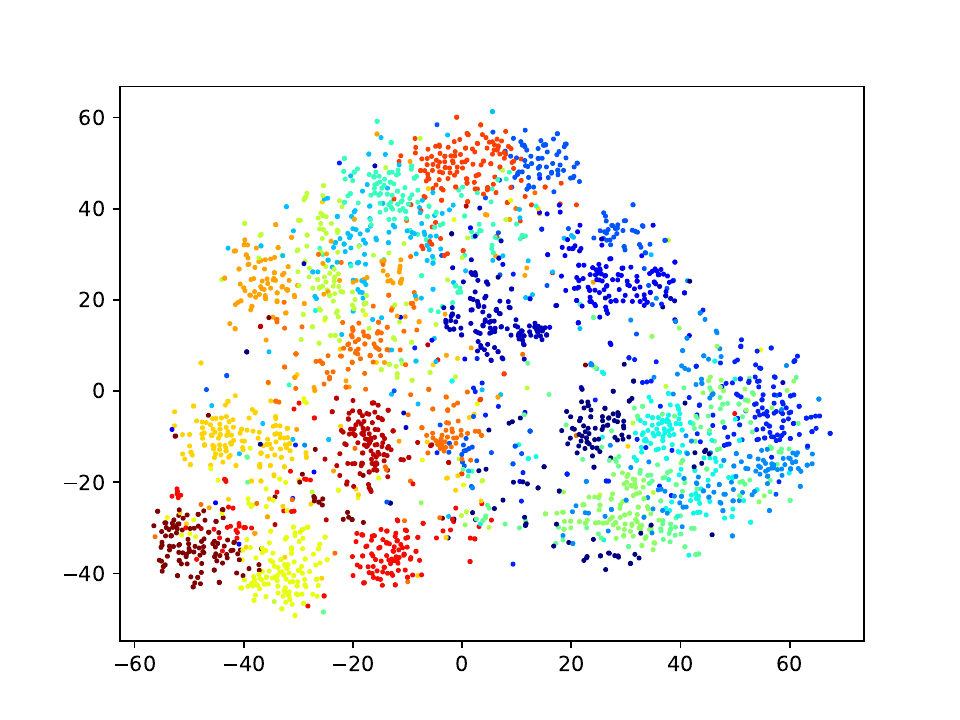}
		\end{minipage}
	}
	\subfloat[5-way 1-shot]{
		\begin{minipage}{0.45\linewidth}
			\centering
			\includegraphics[width=0.9\textwidth]{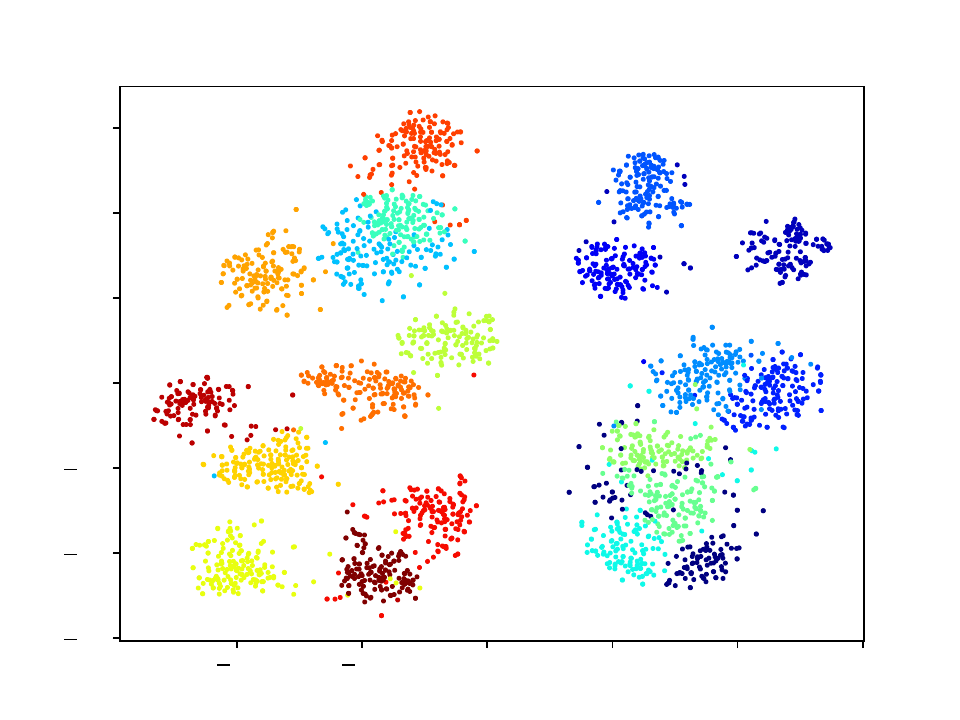}
		\end{minipage}
	}
        \\
        \subfloat[5-way 5-shot ]{
		\begin{minipage}{0.45\linewidth}
			\centering
			\includegraphics[width=0.9\textwidth]{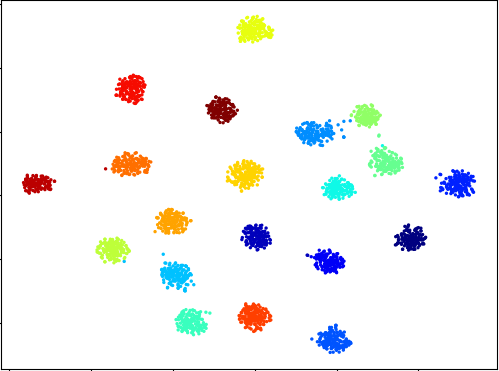}
		\end{minipage}
	}
	\subfloat[5-way 5-shot]{
		\begin{minipage}{0.45\linewidth}
			\centering
			\includegraphics[width=0.9\textwidth]{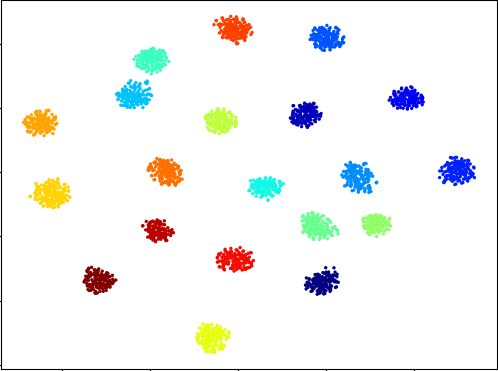}
		\end{minipage}
	}
	\caption{t-SNE visualization of the classification weights for all novel categories in Mini-ImageNet. (a) 1-shot visual-based classifier. (b) 1-shot multi-modal based classifier. (c) 5-shot visual-based classifier. (d) 5-shot multi-modal based classifier.}
	\label{fig:tsne}
\end{figure}

\paragraph{Effect of Multi-modal Knowledge Fusion.} As shown in Figure~\ref{fig:tsne}, we utilize t-SNE visualization to present classifier weights for all novel categories of MiniImageNet. Figure~\ref{fig:tsne}(a) illustrates classifier weights derived solely from visual data in the 1-shot setting, revealing loosely defined clusters with significant category overlap. Conversely, Figure~\ref{fig:tsne}(b) shows the 1-shot results with fused visual and semantic knowledge, where clusters are more compact and distinct. This highlights the substantial benefit of semantic knowledge in scenarios with limited samples. Figure~\ref{fig:tsne}(c) shows the 5-shot classifier weights based solely on visual data, showing more tightly grouped clusters with less overlap. Figure~\ref{fig:tsne}(d) presents the 5-shot results with multi-modal knowledge fusion, where clusters are even more distinct and compact. This demonstrates that rich knowledge still improves FSL performance even with a higher number of samples.

\section{Conclusion}
In this paper, we delve into the previously unexplored potential of harnessing the extensive knowledge available in large multimodal models to empower the pre-trained few-shot learner. As a result, we propose a Synergistic Knowledge Transfer (\textsc{SynTrans}) framework that effectively transfers diverse and complementary knowledge from both visual- and large-language models to address FSL tasks. 
The essence of \textsc{SynTrans} lies in its proficient capability to distill and transform explict visual knowledge and implicit semantic knowledge from these large models into practical classifier weights, thereby significantly improving FSL performance through a multimodal knowledge fusion manner. Experimental results on four benchmark datasets demonstrate the superior efficacy of \textsc{SynTrans} compared to the state-of-the-art methods. 

\section*{Acknowledgments}
This work is supported by the Shenzhen-Hong Kong-Macao Science and Technology Plan Project (Category C) under the Shenzhen Municipal Science and Technology Innovation Commission (Project No. SGDX20230821092359002), and a grant for Collaborative Research with World-leading Research Groups of The Hong Kong Polytechnic University (Project No. G-SACF). Additional support is provided by the Guangdong Natural Science Funds for Distinguished Young Scholars (Grant No. 2023B1515020097), the National Research Foundation, Singapore under its AI Singapore Programme (AISG Award No: AISG3-GV-2023-011), and the Lee Kong Chian Fellowships.

\bibliographystyle{named}
\bibliography{ijcai25}

\end{document}